\pdfoutput=1

\documentclass[11pt]{article}

\usepackage[final]{acl}

\usepackage{times}
\usepackage{latexsym}
\usepackage{tcolorbox}
\tcbuselibrary{skins, breakable, theorems}
\usepackage{booktabs}
\usepackage{multirow}
\usepackage{adjustbox}
\usepackage{xcolor}
\usepackage{soul}
\usepackage{graphicx}
\usepackage{subcaption}
\usepackage{caption}
\usepackage{makecell}
\usepackage{amssymb}
\usepackage{pifont}
\usepackage{algorithm}
\usepackage{algpseudocode}
\usepackage{algorithmicx}

\usepackage[table]{xcolor}
\usepackage[T1]{fontenc}

\usepackage[utf8]{inputenc}

\usepackage{microtype}

\usepackage{inconsolata}

\usepackage{graphicx}

\definecolor{crosscolor}{rgb}{0.969,0.580,0.114} %
\definecolor{checkcolor}{rgb}{0.485,0.640,0.204} %

\def\cmark{{\color{checkcolor}\ding{52}}}
\def\xmark{{\color{crosscolor}\ding{56}}}

\definecolor{MyBlue}{HTML}{a9d5ee}
\definecolor{MyLightBlue}{HTML}{DAEEFA}
\newcommand{\textcolorblue}[1]{
  \begingroup
  \sethlcolor{MyBlue}
  \textcolor{black}{\hl{#1}}
  \endgroup
}
\newcommand{\textcolorlblue}[1]{
  \begingroup
  \sethlcolor{MyLightBlue}
  \textcolor{black}{\hl{#1}}
  \endgroup
}

\title{Meta-Moderator: Empowering Multi-Agent Debate with Meta-Cognition}

\author{
    \textbf{Wentao Hu}$^\spadesuit$\quad \textbf{Zhuoyue WAN}$^\spadesuit$\quad \textbf{Jinhao Shen}$^\spadesuit$ \\
    \textbf{Chen Jason Zhang}$^\spadesuit$\quad \textbf{Xiaoyong Wei}$^{\heartsuit, \spadesuit,}$\thanks{\ Corresponding author}\quad \textbf{Qing Li}$^\spadesuit$ \\
    $^\spadesuit$The Hong Kong Polytechnic University \quad $^\heartsuit$Sichuan University\\
    \texttt{wayne-wt.hu@connect.polyu.hk}\\
    \texttt{\{jason-c.zhang, cs007.wei\}@polyu.edu.hk}
}

\begin{document}
\maketitle

\begin{abstract}

Multi-agent debate can improve large language model reasoning by eliciting diverse hypotheses and critiques, yet its performance is often constrained by weak moderation. Common pipelines rely on fixed budgets, agreement-based stopping, or untrained judges, leading to redundant deliberation and unreliable evidence aggregation.
We cast moderation as a meta-cognitive process, monitoring debate utility, controlling deliberation, and adjudicating a final answer, and introduce \textbf{Meta-Moderator}, a learnable framework that dynamically regulates debate and decides when to finalize an answer.
Meta-Moderator is trained independently of the debaters via outcome-driven policy optimization, making debate regulation an explicit capability rather than an incidental effect of prompting.
Across five benchmarks, Meta-Moderator outperforms widely used decision layers and transfers across tasks and system configurations. Further analyses show that it allocates debate more selectively and reduces mis-aggregation after informative hypotheses appear.\footnote{\ Our code available at \url{https://github.com/Huenao/Meta-Moderator}.}

\end{abstract}
\section{Introduction}

Large language models (LLMs) often achieve stronger reasoning when multiple agents are allowed to interact, critique one another, and iteratively refine their answers \cite{du2023improving, liang-etal-2024-encouraging, chan2024chateval}. This multi-agent debate (MAD) paradigm has produced consistent gains across tasks such as mathematical reasoning, multi-hop question answering, and factual verification, motivating the common belief that structured interaction yields more reliable reasoning than a single agent \cite{zhang-xiong-2025-debate4math, hu-etal-2025-removal}.

\begin{figure}[t]
    \centering
    \includegraphics[width=0.99\linewidth]{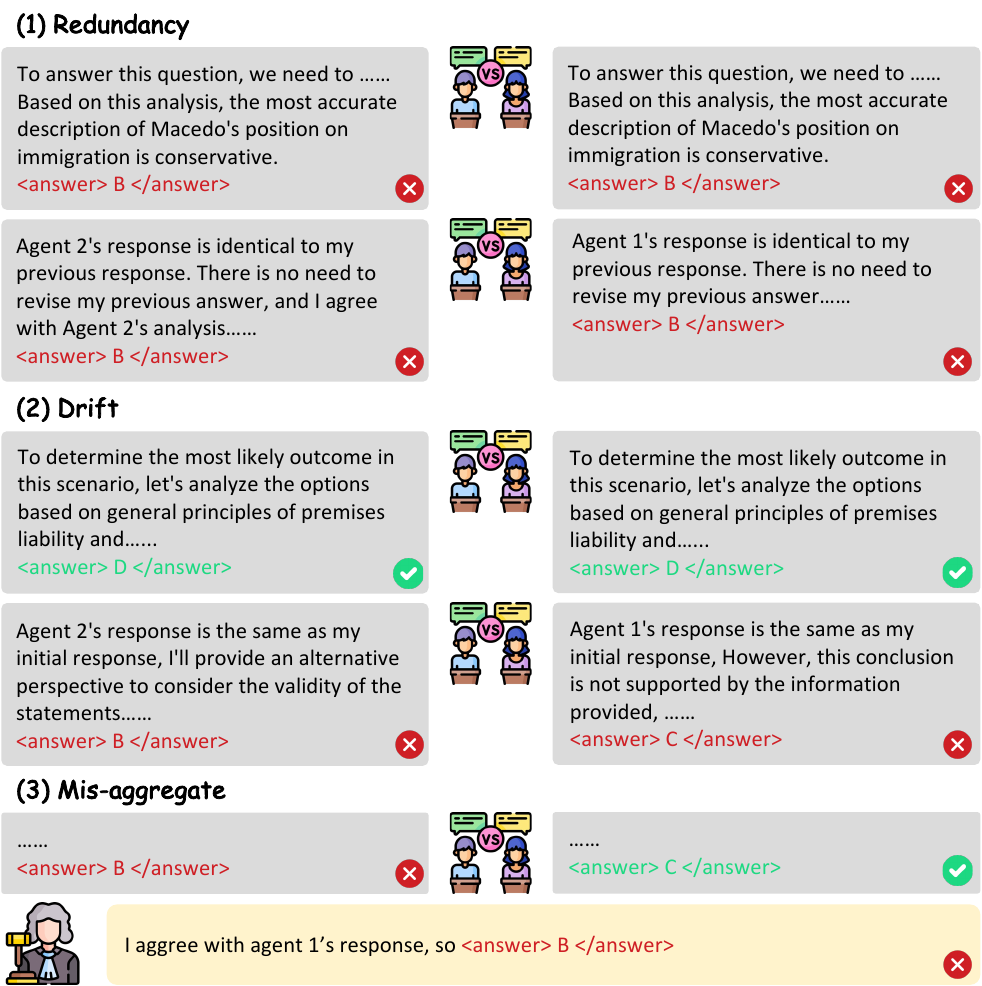}
    \caption{Three common issues in standard multi-agent debate: redundancy, drift, and mis-aggregation.}
    \label{fig:intro}
\end{figure}

Despite its promise, MAD is far from reliably beneficial. As shown in Figure~\ref{fig:intro}, debates can (i) become \textbf{redundant}, expending additional rounds that contribute little new information; (ii) \textbf{drift} away from the key uncertainty, accumulating plausible but irrelevant details; or (iii) \textbf{mis-aggregate} high-quality arguments into an incorrect final decision, for instance due to superficial consensus, last-round bias, or voting failures under dispersed opinions \cite{becker2024multi, zheng2024picture, choi2025debate, cui2025free, hong2025qualbenchbenchmarkingchinesellms}. These failure modes arise naturally when collective deliberation is run without reliable mechanisms for assessing progress, allocating deliberation, and committing to a final decision.

We argue that these issues point to a central, under-modeled bottleneck in MAD: the \emph{moderator}. In real-world adversarial deliberation (e.g., courtroom argumentation), advocates focus on local evidence and persuasive reasoning, while a neutral adjudicator maintains global context, tracks what has been resolved, and decides when the discussion is ready for a decision. By contrast, most MAD systems treat moderation as a heuristic add-on: interaction is governed by fixed budgets, shallow consensus checks, or voting-based aggregation \cite{choi2025debate, cui2025free}. Recent work has begun to explore measurement-driven controllers that stop when debate signals plateau or when judge dynamics stabilize \cite{chang2025multi, hu2025multiagent}. However, these approaches still rely on predefined signals and hand-chosen stopping criteria, rather than learning a moderation policy optimized for end-task accuracy. Some approaches introduce an LLM judge, but it typically operates as a prompted black-box evaluator rather than a policy optimized for regulating debate dynamics \cite{liang-etal-2024-encouraging, hong2026uxbenchbenchmarkinguserexperience}. As a result, the very capabilities that determine whether debate translates into accuracy gains are left implicit and fragile.

This motivates a meta-cognitive view of moderation. Drawing inspiration from meta-reasoning and metacognition, which study how cognitive processes are monitored and regulated under limited resources \cite{ackerman2017meta, ball2025metareasoning, gao2024meta, de2024rational}, we characterize effective moderation as a tightly coupled loop of three functions: \textbf{Monitoring} whether deliberation is making substantive progress versus repeating or drifting; \textbf{Control} allocating deliberation by deciding whether to prolong deliberation or commit to a decision \cite{wei_coaching}; and \textbf{Adjudication} synthesizing a final answer from the debate content in a way that is robust to superficial agreement and aggregation biases. Under this perspective, moderation is not merely another round of object-level reasoning; it is a distinct, policy-like capability that must decide \emph{when} further deliberation is worthwhile and \emph{when} to commit to a final decision.

Motivated by this framing, we introduce \textbf{Meta-Moderator}, a framework for learnable moderation that treats debate regulation as a meta-level decision problem. Meta-Moderator learns an explicit moderation policy that decides, at each round, whether further deliberation is warranted or whether to stop and commit to a final answer. We train this policy independently from debaters with outcome-driven reinforcement learning, so that moderation becomes a dedicated, learnable competence rather than an incidental outcome of prompting or heuristic aggregation.

Our contributions are threefold:
\begin{itemize}
    \item We identify moderation as a central bottleneck in multi-agent debate and formalize MAD as a meta-level regulation problem characterized by a monitoring-control-adjudication loop.
    \item We propose \textbf{Meta-Moderator}, a framework for \emph{learnable moderation} that instantiates an explicit moderation policy to decide whether to continue deliberation or stop and commit to a final answer.
    \item Across multiple reasoning tasks, we show that learned meta-level regulation improves final answer accuracy compared to common MAD baselines.
\end{itemize}
\section{Related Work}

\subsection{Multi-Agent Debate}
Multi-agent debate (MAD) aims to improve LLM reasoning by having agents propose answers, critique one another, and refine arguments \cite{du2023improving}. Subsequent work promotes divergent thinking, structured argumentation, and stronger verification in multimodal and retrieval-augmented settings \cite{liang-etal-2024-encouraging, zheng2024picture, hu-etal-2025-removal}.

However, recent analyses suggest that debate is not consistently beneficial. Gains are often attributable to diversity in initial answers rather than multi-round deliberation, and debate trajectories do not reliably increase correctness \cite{choi2025debate}. Other work further reports conformity and bias amplification, where consensus-seeking dynamics suppress minority hypotheses and propagate shared errors \cite{cui2025free}. These findings imply that debate effectiveness depends not only on debaters’ object-level reasoning, but also on how debate is monitored, budgeted, and aggregated into a final decision.

Most MAD pipelines rely on fixed debate lengths, agreement-based stopping, or untrained judges. Measurement-driven stopping moves beyond fixed rounds but still depends on predefined signals and hand-tuned criteria \cite{chang2025multi, hu2025multiagent}. We address this gap by learning a moderation policy optimized for end-task accuracy.

\subsection{Metacognition for LLMs}
Recent work on metacognition in LLMs mainly focuses on self-monitoring and self-regulation \cite{gao2024meta}. Typical operationalizations include confidence estimation, uncertainty awareness, and failure prediction \cite{ji-an2025language}. Although such self-evaluative signals can be elicited, they are often miscalibrated and only weakly aligned with models' true knowledge limitations \cite{wang2025decoup, wang-zhao-2024-metacognitive}. Complementary work frames meta-reasoning as a control mechanism over actions such as tool use, retrieval, or allocating additional computation \cite{li-etal-2025-adaptive, alazraki-rei-2025-meta, de2024rational}.

In contrast, we study metacognition as coordination in multi-agent deliberation: a moderator monitors agents' externalized reasoning traces and regulates whether further debate is worthwhile. This motivates a learnable moderation policy for MAD.

\subsection{Reinforcement Learning for LLMs}
Reinforcement learning (RL) is increasingly used to shape LLM behavior beyond supervised fine-tuning \cite{wu-etal-2025-ddxtutor, wan-DataVisT5}, including improving multi-step reasoning and learning decision policies for LLM-based agents \cite{shao2024deepseekmath, guo2025deepseek, chen2026syncloopmultimodaldualloopframework}. RL has also been applied to higher-level control, such as planning, tool selection, and allocating search or computation, making it a natural approach for learning meta-level policies over hierarchical reasoning processes \cite{jin2025search, chen2025research, qian2025toolrl, wu2025structure}.

However, prior work largely optimizes a single agent's reasoning or tool-use policy, leaving the moderation policy, when to stop and how to adjudicate multi-agent evidence, underexplored. We address this gap by training a meta-level moderator with GRPO to directly optimize end-task accuracy.

\section{Meta-Moderator}

\subsection{Problem Formulation}
Let $\mathcal{X}$ denote the input space (e.g., questions) and $\mathcal{Y}$ the output space (e.g., free-form or multiple-choice answers). 
Given an instance $x \in \mathcal{X}$ with gold label $y^\star \in \mathcal{Y}$, we consider a multi-agent debate system composed of $N$ LLM-based debaters 
$\mathcal{D}=\{d_1,\dots,d_N\}$.

\paragraph{Agent responses.}
Each debater produces a textual response at round $t$, denoted by $r_{d_i}^t$. 
In our setting, $r_{d_i}^t$ contains both \emph{reasoning} and a \emph{final answer}; we denote the extracted answer by $y_{d_i}^t = f_{\mathrm{ans}}(r_{d_i}^t)$.
At initialization ($t=0$), each debater independently generates an initial response:
\begin{equation}
r_{d_i}^0 \sim {\tt Init}(x), \quad i\in\{1,\dots,N\},
\end{equation}
where ${\tt Init}$ samples an initial response conditioned on $x$.

\paragraph{Debate dynamics.}
The debate runs for at most $T_{\max}$ rounds. At each round $t\ge 1$, debater $d_i$ observes the set of other debaters' responses from the previous round,
\begin{equation}
\mathcal{R}^{t}_{-d_i} = \{\,r_{d_j}^{t-1}\mid j\in\{1,\dots,N\},\, j\neq i\,\},
\end{equation}
and updates its response via a one-round debate operator ${\tt Debate}$:
\begin{equation}
r_{d_i}^t \sim {\tt Debate}\big(x;\mathcal{R}^{t}_{-d_i}\big).
\end{equation}
All debaters update in parallel using the responses from the previous round.

Let $\mathcal{H}_t$ denote the debate transcript from initialization through round $t$:
\begin{equation}
\mathcal{H}_t = \{ r_{d_i}^\tau \mid i\in\{1,\dots,N\},\; \tau \in \{1,\dots,t\} \}.
\end{equation}

\paragraph{Decision layer.}
A moderator (decision layer) observes the debate and may terminate at any round $t\le T_{\max}$. Formally, the system outputs
\begin{equation}
\hat{y} = \mathcal{F}(\mathcal{H}_t),
\end{equation}
where $t$ is the (data-dependent) stopping time chosen by the moderator. The function $\mathcal{F}$ abstracts the \emph{decision layer} of debate: it can be instantiated as majority voting over $\{y_{d_i}^t\}_{i=1}^N$, or as a judge model conditioning on $\mathcal{H}_t$.

\paragraph{Objective.}
Our goal is to maximize end-task correctness of the overall debate system by improving the decision layer:
\begin{equation}
\max \; \mathbb{E}\big[\mathbf{1}\{\hat{y}=y^\star\}\big],
\end{equation}
where the expectation is over task instances and the stochasticity of debater generation and debate updates.
Next, we will parameterize $\mathcal{F}$ as a \emph{learnable} moderator policy that decides when to stop and how to adjudicate, enabling the system to better exploit debate evidence when producing $\hat{y}$.

\subsection{Meta-Moderator Policy}
\label{sec:meta_moderator_policy}

\begin{figure*}
    \centering
    \includegraphics[width=0.95\linewidth]{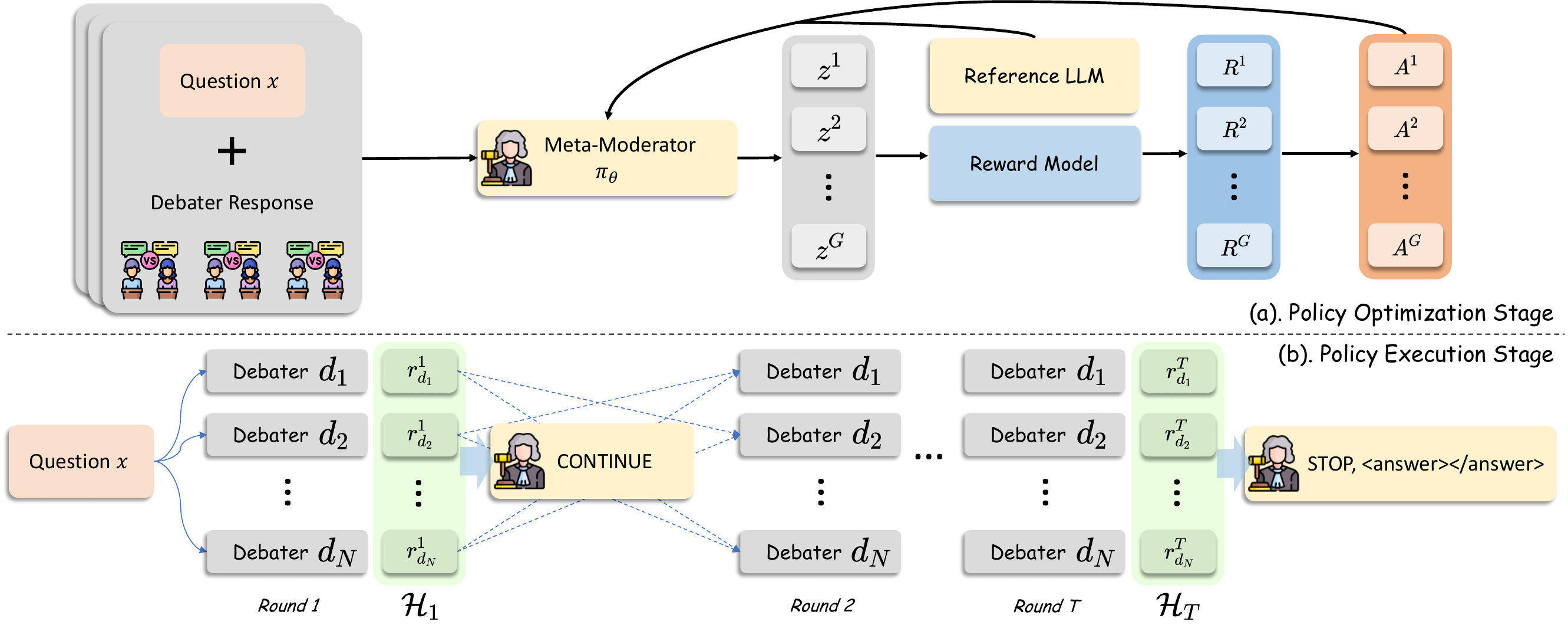}
    \caption{Overview of Meta-Moderator. The moderator monitors debate states and adjudicates a final answer when the debate stops; it is trained via outcome-driven policy optimization and used as a drop-in decision layer.}
    \label{fig:framework}
\end{figure*}

The formulation above separates the debate layer, which generates multi-agent evidence through interaction, from the decision layer $\mathcal{F}$, which maps debate evidence to a single prediction and an adaptive stopping point. Most existing MAD systems instantiate $\mathcal{F}$ with static heuristics or a prompted judge. In contrast, we parameterize $\mathcal{F}$ as a \textbf{Meta-Moderator}: a learnable meta-level policy that performs monitoring, control (adaptive stopping), and adjudication over the evolving debate.

\paragraph{Moderator state.}
At round $t$, the moderator observes a debate state
\begin{equation}
s_t = \phi(\mathcal{H}_t),
\end{equation}
where $\phi(\cdot)$ specifies the moderator's input view. In general, $\phi$ may expose the full transcript or a compressed representation. In this work, we use a round-level view:
\begin{equation}
s_t = \phi\left(x, \{r_{d_i}^{t}\}_{i=1}^N\right),
\end{equation}
constructed from the question and the current-round debater responses. Importantly, debaters' responses at round $t$ may include summaries of prior discussion, allowing $s_t$ to encode relevant history without explicitly concatenating $\mathcal{H}_t$.

\paragraph{Moderator policy and structured output.}
The moderator is implemented as a conditional language-model policy $\pi_\theta(\cdot\mid s_t)$ that produces a structured moderation output.

\paragraph{Monitoring.}
Monitoring refers to assessing whether the current debate state contains sufficient, non-redundant evidence for a reliable decision. 
Rather than introducing an explicit scalar monitor, we treat monitoring as an implicit internal signal represented by the policy when producing its control decision at round $t$.
Formally, the control distribution $\pi_\theta(a_t\mid s_t)$ depends on $s_t$ and thus implicitly captures whether additional deliberation is expected to be beneficial under the current debate evidence.

\paragraph{Control.}
Control uses the monitoring assessment to regulate deliberation depth. The moderator selects a control decision
\begin{equation}
a_t\in\{\texttt{CONTINUE},\texttt{STOP}\},
\end{equation}
where \texttt{CONTINUE} allocates an additional debate round and \texttt{STOP} terminates deliberation.
While moderation can in principle include richer interventions that shape the next-step information flow (e.g., requesting counterexamples or verification), we focus on this minimal control space to isolate the core question: \emph{is further interaction beneficial given $s_t$?}

\paragraph{Adjudication.}
When the moderator decides to stop, it produces a single final prediction by aggregating multi-agent evidence in the current state. This is realized as conditional generation under the same policy:
\begin{equation}
\hat{y}_t \sim \pi_\theta(\cdot \mid s_t, a_t=\texttt{STOP}),
\end{equation}
yielding a learnable instantiation of $\mathcal{F}$ that may adjudicate at an intermediate round rather than relying on a fixed horizon.
Unlike surface-level consensus, learnable adjudication can weigh competing rationales and identify unresolved inconsistencies before committing to $\hat{y}_t$.

Overall, Meta-Moderator performs inter-agent meta-level regulation by monitoring debate utility, controlling whether to continue deliberation, and adjudicating a final answer from accumulated debate evidence.

\subsection{Training Data Construction}
We train Meta-Moderator from an offline dataset of debate states.
For each training instance $x \in \mathcal{X}'$, we run a fixed multi-agent debate protocol to obtain a multi-round transcript $\mathcal{H}_t$ up to $T_{\max}$.
We then extract, for each round $t$, the moderator-observable state
\begin{equation}
s_t = \phi\!\left(x, \{r_{d_i}^{t} \mid i\in\{1,\dots,N\}\}\right),
\end{equation}
where $\phi(\cdot)$ is a deterministic formatting function that maps the question and the current-round debater responses into the moderator's input.

Each training example is a tuple $(s_t, y^\star, \eta_t)$.
Here $\eta_t \in \{0,1\}$ is an offline auxiliary signal indicating whether a correct
candidate answer already appears among debaters at round $t$.
We compute $\eta_t$ by automatically extracting each debater's predicted answer from $r_{d_i}^{t}$,
followed by dataset-specific normalization and matching against $y^\star$.
This signal is used only to shape rewards during training; at inference time, Meta-Moderator observes only $s_t$ and never accesses gold labels.

\subsection{Reinforcing Meta-Cognition in Debate}
\label{sec:meta_cog}

We train the Meta-Moderator to exhibit \emph{meta-cognition} over multi-agent debate:
it must \emph{monitor} whether the current debate state already contains sufficient evidence for a correct answer,
and \emph{control} the process by either continuing deliberation or committing to a final answer.
Given a debate state $s_t$, the moderator generates a textual decision
$z_t \sim \pi_{\theta}(\cdot \mid s_t)$.
We constrain the output space to two executable forms:
\begin{equation}
z_t \in 
\Big\{\texttt{CONTINUE},\;
\texttt{STOP}\,\Vert\,\langle\texttt{answer}\rangle \hat{y}_t \langle/\texttt{answer}\rangle \Big\},
\end{equation}
which is parsed into an action-answer pair $(a_t,\hat{y}_t)$.

\paragraph{Reward.}
We use three additive reward components to reinforce (i) protocol-compliant decisions,
(ii) sufficiency monitoring for debate termination, and (iii) correctness upon commitment:
\begin{equation}
\begin{aligned}
R(s_t, z_t) = R_{\mathrm{fmt}}(z_t)&+ R_{\mathrm{ctrl}}(s_t, a_t) 
\\ & + R_{\mathrm{ans}}(y^\star, a_t, \hat{y}_t).
\end{aligned}
\end{equation}

\noindent
\textbf{Format reward} $R_{\mathrm{fmt}}$ enforces an \emph{observable} and machine-parseable control protocol:
$\texttt{CONTINUE}$ alone, or $\texttt{STOP}$ with a properly delimited $\langle\texttt{answer}\rangle$ span.
We assign $1$ for a valid format; for $\texttt{STOP}$ without an answer span we give partial credit ($0.5$); otherwise $0$.

\noindent
\textbf{Control reward} $R_{\mathrm{ctrl}}$ reinforces \emph{sufficiency monitoring}:
the moderator should stop when the debate already contains a gold-matching proposal, and continue otherwise.
We use an offline indicator $\eta_t \in \{0,1\}$, where $\eta_t=1$ means that at round $t$
at least one debater proposes a gold-matching answer (estimated by automatic answer extraction and normalization).
We reward $\texttt{STOP}$ when $\eta_t=1$, and reward $\texttt{CONTINUE}$ when $\eta_t=0$; otherwise the reward is $0$.
This gold-derived signal is used only for offline reward shaping during training and is never provided as an inference-time input.

\noindent
\textbf{Answer reward} $R_{\mathrm{ans}}$ reinforces \emph{commitment correctness} and is only applied at termination:
\begin{equation}
R_{\mathrm{ans}}(y^\star, a_t, \hat{y}_t)=
\begin{cases}
\mathbb{I}[\hat{y}_t = y^\star], & a_t=\texttt{STOP},\\
0, & a_t=\texttt{CONTINUE}.
\end{cases}
\end{equation}
Dataset-specific matching is used for correctness.

\paragraph{GRPO optimization.}
For each state $s_t$, we sample $G$ candidate decisions $\{z_t^{(i)}\}_{i=1}^{G}$ from the old policy
and compute total rewards $\{R^{(i)}\}_{i=1}^{G}$.
We construct group-relative advantages by normalizing rewards within the group:
\begin{equation}
A^{(i)}=\frac{R^{(i)}-\mu_{\{R\}}}{\sigma_{\{R\}}},
\end{equation}
and update $\pi_\theta$ with a clipped surrogate objective and a KL regularizer, following GRPO \cite{shao2024deepseekmath}.
More details are presented in Appendix~\ref{app:learning}.

\subsection{Inference-Time Moderation Protocol}
At test time, Meta-Moderator is used as a drop-in regulator for any fixed set of debaters.
Given an instance $x$, the system iterates rounds $t=0,1,\dots,T_{\max}$:
(i) debaters produce responses $\{r_{d_i}^{t}\}_{i=1..N}$ under the debate protocol;
(ii) the moderator state $s_t=\phi(x,\{r_{d_i}^{t}\}_{i=1..N})$ is constructed;
(iii) the moderator samples or decodes a decision $z_t$ from $\pi_{\theta}(\cdot\mid s_t)$ and parses it into $(a_t,\hat{y}_t)$.
If $a_t=\texttt{CONTINUE}$, the debate proceeds to the next round.
If $a_t=\texttt{STOP}$, the system terminates and returns $\hat{y}_t$ as the final answer.
If no stop decision is made by $T_{\max}$, the system terminates at $T_{\max}$ using the moderator's final output.
\begin{table*}[t]
    \centering
    \adjustbox{max width=0.95\textwidth}{
    \begin{tabular}{lcccccc} 
        \toprule
        Method & Use Moderator & GSM8K & AMC & MATH500 & StrategyQA & MMLU \\ 
        \midrule
        \rowcolor{gray!15}\textbf{\textit{Single-agent}} & \multicolumn{6}{c}{\textit{Llama-3.1-8B-Instruct}} \\ 
        Naive & - & 22.20 & 7.23 & 3.80 & 68.20 & 57.00 \\
        Reflection \cite{shinn2023reflexion} & - & 30.40 & 3.61 & 7.40 & 57.80 & 34.20 \\
        CoT \cite{wei2022chain} & - & \colorbox{MyLightBlue}{\textbf{83.40}} & 16.87 & 33.60 & 67.40 & 62.60 \\
        \rowcolor{gray!15}\textbf{\textit{Multi-Agent Debate}} & \multicolumn{6}{c}{\textit{Llama-3.1-8B-Instruct}} \\
        + Majority-Voting \cite{cui2025free} & \xmark & 79.00 & 25.20 & 27.80 & 66.40 & 61.80 \\
        + Consensus \cite{du2023improving} & \xmark & 81.80 & 22.89 & 27.20 & \colorbox{MyLightBlue}{\textbf{69.80}} & \colorbox{MyLightBlue}{\textbf{64.00}} \\
        + LLM-as-Judge \cite{liang-etal-2024-encouraging} & \cmark & 80.00 & \colorbox{MyLightBlue}{\textbf{25.30}} & 33.60 & 63.00 & 62.00 \\
        + \textbf{Meta-Moderator$^*$ (Ours)} & \cmark & 79.80 & 19.28 & \colorbox{MyLightBlue}{\textbf{33.80}} & 68.80 & 61.60 \\
        + \textbf{Meta-Moderator (Ours)} & \cmark & \colorbox{MyBlue}{\textbf{83.80}} & \colorbox{MyBlue}{\textbf{27.71}} & \colorbox{MyBlue}{\textbf{35.80}} & \colorbox{MyBlue}{\textbf{72.00}} & \colorbox{MyBlue}{\textbf{67.20}} \\
        \midrule
        \rowcolor{gray!15}\textbf{\textit{Single-agent}} & \multicolumn{6}{c}{\textit{Qwen-2.5-7B-Instruct}} \\ 
        Naive & - & 20.60 & 15.66 & 4.00 & 65.20 & 69.60 \\
        Reflection \cite{shinn2023reflexion} & - & 89.20 & 20.48 & 18.80 & 69.80 & 68.60 \\
        CoT \cite{wei2022chain} & - & 90.20 & 21.69 & 15.40 & \colorbox{MyLightBlue}{\textbf{70.00}} & 61.20 \\
        \rowcolor{gray!15}\textbf{\textit{Multi-Agent Debate}} & \multicolumn{6}{c}{\textit{Qwen-2.5-7B-Instruct}} \\
        + Majority-Voting \cite{cui2025free} & \xmark & 90.80 & 26.51 & 14.40 & 67.20 & 68.80 \\
        + Consensus \cite{du2023improving} & \xmark & 90.60 & 26.51 & 14.40 & 67.20 & 68.80 \\
        + LLM-as-Judge \cite{liang-etal-2024-encouraging} & \cmark & 90.80 & \colorbox{MyBlue}{\textbf{43.37}} & \colorbox{MyLightBlue}{\textbf{19.00}} & 67.40 & 69.80 \\
        + \textbf{Meta-Moderator$^*$ (Ours)} & \cmark & \colorbox{MyLightBlue}{\textbf{90.80}} & 39.76 & 13.80 & 67.60 & \colorbox{MyBlue}{\textbf{70.60}} \\
        + \textbf{Meta-Moderator (Ours)} & \cmark & \colorbox{MyBlue}{\textbf{91.20}} & \colorbox{MyLightBlue}{\textbf{42.96}} & \colorbox{MyBlue}{\textbf{29.40}} & \colorbox{MyBlue}{\textbf{70.40}} & \colorbox{MyLightBlue}{\textbf{69.80}} \\
        \bottomrule
    \end{tabular}
    }
    \caption{Overall results of \textbf{Meta-Moderator} and baselines on five benchmarks under two backbone settings.\textbf{Meta-Moderator$^\ast$} denotes the \emph{untrained} (prompted) moderator. \textcolorblue{Blue} and \textcolorlblue{light blue} denote the best and second-best performances, respectively. All methods are evaluated under the same debate protocol and decoding configuration.}
    \label{tab:main}
\end{table*}

\section{Experiments}

\subsection{Experimental Settings}
\paragraph{Baselines.}
We compare \textbf{Meta-Moderator} with single-agent prompting and multi-agent debate baselines under a unified setup.
For single-agent methods, we include \textbf{Naive}, \textbf{CoT} \cite{wei2022chain}, and \textbf{Reflection} \cite{shinn2023reflexion}.
For MAD, we use a standard protocol where debaters iteratively revise responses conditioned on peers' previous-round outputs \cite{du2023improving}, and vary only the decision layer: \textbf{Majority Voting} \cite{cui2025free}, \textbf{Consensus} \cite{du2023improving}, or \textbf{LLM-as-a-Judge} \cite{liang-etal-2024-encouraging}.

\paragraph{Datasets and Metrics.}
We evaluate on five benchmarks covering mathematical reasoning, logical reasoning, and general knowledge: \textbf{GSM8K} \cite{cobbe2021training}, \textbf{AMC}, \textbf{MATH500} \cite{hendrycks2021measuring}, \textbf{StrategyQA} \cite{geva-etal-2021-aristotle}, and \textbf{MMLU} \cite{hendrycks2020measuring}.
For large test sets, we randomly sample 500 instances; for smaller benchmarks, we use the full test set (e.g., \textbf{AMC} has 83 instances). We report \textbf{accuracy}.
To train Meta-Moderator, we sample 500 instances each from the training splits of \textbf{GSM8K} and \textbf{MMLU} to construct per-round moderation decisions.

\paragraph{Implementation Details.}
We instantiate both debaters and the moderator with instruction-tuned LLMs, using Llama-3.1-8B-Instruct and Qwen-2.5-7B-Instruct as backbones \cite{dubey2024llama, bai2023qwen}.
Unless otherwise specified, we use $N{=}2$ debaters and a maximum debate budget of $T_{\max}{=}5$ rounds.
All methods use the same debater protocol and decoding configuration; only the decision layer differs.
Additional backbone settings are reported in Appendix~\ref{app:extra_results}.

\subsection{Main Results}

\begin{table*}[t]
    \centering
    \adjustbox{max width=0.95\textwidth}{
    \begin{tabular}{lccccc}
        \toprule
        Method & GSM8K & AMC & MATH500 & StrategyQA & MMLU \\
        \midrule
        \rowcolor{gray!15}
        \textbf{\textit{Multi-Agent Debate}} & \multicolumn{5}{c}{\textit{Llama-3.1-8B-Instruct} \& \textit{Llama-3.2-3B-Instruct}} \\
        + LLM-as-Judge \cite{liang-etal-2024-encouraging} & \colorbox{MyLightBlue}{\textbf{79.80}} & \colorbox{MyBlue}{\textbf{26.51}} & \colorbox{MyBlue}{\textbf{29.60}} & \colorbox{MyLightBlue}{\textbf{64.80}} & \colorbox{MyLightBlue}{\textbf{62.80}} \\
        + \textbf{Meta-Moderator$^*$ (Ours)} & 79.40 & \colorbox{MyBlue}{\textbf{26.51}} & 27.80 & 60.60 & 61.40 \\
        + \textbf{Meta-Moderator (Ours)} & \colorbox{MyBlue}{\textbf{83.60}} & 22.89 & \colorbox{MyLightBlue}{\textbf{28.00}} & \colorbox{MyBlue}{\textbf{71.20}} & \colorbox{MyBlue}{\textbf{65.60}} \\
        \bottomrule
    \end{tabular}
    }
    \caption{Results with decoupled debater and moderator backbones: debaters use Llama-3.1-8B-Instruct while the moderator uses Llama-3.2-3B-Instruct.}
    \label{tab:generalization}
\end{table*}
\begin{table*}[t]
    \centering
    \adjustbox{max width=0.95\textwidth}{
    \begin{tabular}{lccccc}
        \toprule
        Method & GSM8K & AMC & MATH500 & StrategyQA & MMLU \\
        \midrule
        \rowcolor{gray!15}\textbf{\textit{Multi-Agent Debate}} & \multicolumn{5}{c}{\textit{Llama-3.1-8B-Instruct}} \\
        + Majority-Voting \cite{cui2025free} & 77.40 & 18.07 & 23.60 & 68.20 & 62.80 \\
        + Consensus \cite{du2023improving} & \colorbox{MyLightBlue}{\textbf{79.00}} & 20.48 & 23.80 & \colorbox{MyLightBlue}{\textbf{72.00}} & 63.40 \\
        + LLM-as-Judge \cite{liang-etal-2024-encouraging} & 78.20 & \colorbox{MyBlue}{\textbf{24.10}} & \colorbox{MyLightBlue}{\textbf{31.60}} & 69.60 & \colorbox{MyLightBlue}{\textbf{63.80}} \\
        + \textbf{Meta-Moderator$^*$ (Ours)} & 78.40 & \colorbox{MyBlue}{\textbf{24.10}} & 29.40 & 67.80 & 63.60 \\
        + \textbf{Meta-Moderator (Ours)} & \colorbox{MyBlue}{\textbf{83.80}} & 22.89 & \colorbox{MyBlue}{\textbf{35.60}} & \colorbox{MyBlue}{\textbf{72.60}} & \colorbox{MyBlue}{\textbf{66.00}} \\
        \bottomrule
    \end{tabular}
    }
    \caption{Results with more debaters ($N{=}3$) under the same debate budget.}
    \label{tab:debaters_num}
\end{table*}
\begin{table*}[t]
    \centering
    \adjustbox{max width=0.95\textwidth}{
    \begin{tabular}{lccccc}
        \toprule
        Method & GSM8K & AMC & MATH500 & StrategyQA & MMLU \\
        \midrule
        \rowcolor{gray!15}\textbf{\textit{Multi-Agent Debate}} & \multicolumn{5}{c}{\textit{Llama-3.1-8B-Instruct}} \\
        + \textbf{Meta-Moderator$^*$ (Ours)} & 79.80 & 19.28 & 33.80 & 68.80 & 61.60 \\
        + \textbf{Meta-Moderator (Ours)} & \colorbox{MyBlue}{\textbf{83.80}} & \colorbox{MyBlue}{\textbf{27.71}} & \colorbox{MyBlue}{\textbf{35.80}} & \colorbox{MyBlue}{\textbf{72.00}} & \colorbox{MyBlue}{\textbf{67.20}} \\
        + Adjudication Only  & 81.80 & 24.10 & 28.00 & 62.00 & 63.60 \\
        + Adaptive Stopping Only (Majority-Voting) & 79.40 & 24.10 & 29.40 & 66.20 & 62.40 \\
        + Adaptive Stopping Only (LLM-as-Judge) & 80.40 & 27.71 & 33.60 & 66.80 & 62.20 \\
        \bottomrule
    \end{tabular}
    }
    \caption{Ablation separating adaptive stopping and learned adjudication under the same debate protocol.}
    \label{tab:stop_adj}
\end{table*}

We evaluate \textbf{Meta-Moderator} against single-agent prompting and multi-agent debate baselines across five benchmarks.
As shown in Table~\ref{tab:main}, Meta-Moderator generally improves over common MAD decision layers and achieves the strongest overall performance across the evaluated settings, supporting the value of \emph{learnable} debate regulation.
By contrast, heuristic strategies (e.g., majority voting and consensus-based stopping) are often brittle and can underperform strong single-agent CoT prompting, indicating that additional interaction alone does not reliably yield accuracy gains.

Meta-Moderator is particularly beneficial where heuristic aggregation is brittle, including StrategyQA, MATH500, and MMLU, while prompted LLM-as-a-judge remains competitive on AMC.
Relative to voting-based aggregation and prompted LLM-as-a-judge, Meta-Moderator more reliably converts useful debate evidence into final-answer accuracy, suggesting that learning \emph{when} to stop and \emph{how} to adjudicate helps turn debate into gains.
The learned moderator is also robust across backbone families and model strengths: while heuristic MAD variants can degrade through error propagation and conformity, Meta-Moderator remains competitive with strong single-agent baselines, implying that meta-level regulation can mitigate imperfect object-level reasoning.

Finally, the untrained variant \textbf{Meta-Moderator$^*$} is consistently weaker than the trained moderator and can underperform heuristic alternatives on multiple datasets.
This gap suggests that effective moderation is not merely a byproduct of role prompting; rather, it requires learning a stable policy via outcome-driven optimization.

\subsection{Generalization}
\label{sec:generalization}
We evaluate the generalization of Meta-Moderator along three dimensions.

\paragraph{Cross-task generalization.}
Trained only on GSM8K and MMLU, Meta-Moderator improves accuracy on unseen benchmarks (AMC, MATH500, and StrategyQA; Table~\ref{tab:main}), suggesting that it learns transferable signals of debate utility and decision readiness rather than dataset-specific heuristics.

\paragraph{Cross-backbone generalization.}
To decouple debater and moderator backbones, we pair Llama-3.1-8B-Instruct debaters with a smaller Llama-3.2-3B-Instruct moderator (Table~\ref{tab:generalization}). Meta-Moderator remains effective and outperforms prompted LLM-as-a-Judge on several benchmarks, suggesting that gains come from the learned policy rather than moderator scale.

\begin{table}[t]
    \centering
    \setlength{\tabcolsep}{3pt} 
    \renewcommand{\arraystretch}{1.25}
    
    \resizebox{\columnwidth}{!}{
    \begin{tabular}{lccccc}
        \toprule
        & GSM8K & AMC & MATH & StratQA & MMLU \\ 
        \midrule        
        \multicolumn{6}{l}{\textbf{\textit{Meta-Moderator$^*$}}} \\
        $y^* \in \mathcal{H}_t (\uparrow)$ & \textbf{432} & 21 & 180 & \textbf{386} & \textbf{385} \\
        $\hat{y}\neq y^* \dots (\downarrow)$ & 33 & 8 & 25 & 42 & 77 \\
        \midrule
        \multicolumn{6}{l}{\textbf{\textit{Meta-Moderator}}} \\
        $y^* \in \mathcal{H}_t (\uparrow)$ & 431 & \textbf{24} & \textbf{187} & 369 & 351 \\
        $\hat{y}\neq y^* \dots (\downarrow)$ & \textbf{12} & \textbf{1} & \textbf{8} & \textbf{9} & \textbf{15} \\
        \bottomrule
    \end{tabular}
    }
    \caption{Analysis of debate coverage and adjudication.}
    \label{tab:oracle_single}
\end{table}

\begin{figure*}[h]
    \centering
    \includegraphics[width=0.95\linewidth]{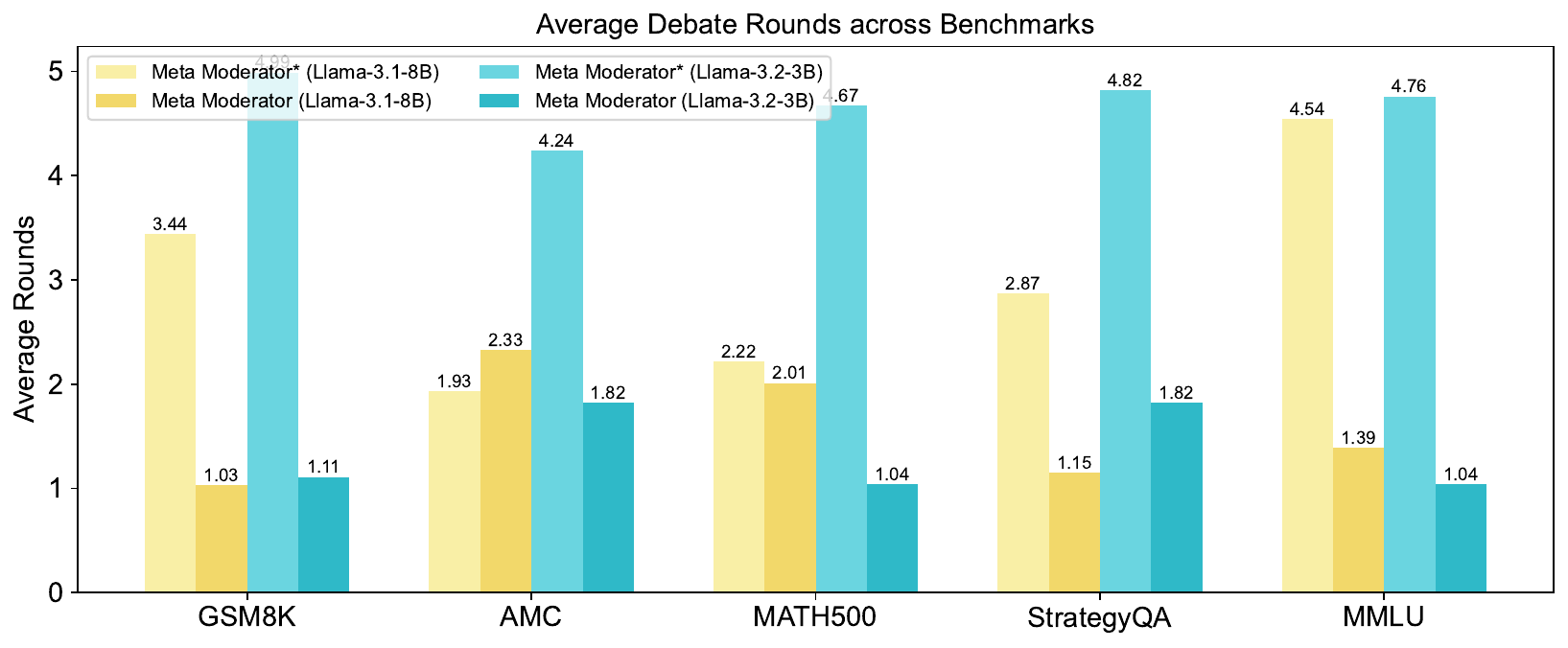}
    \caption{Average debate rounds across benchmarks, comparing the untrained Meta-Moderator$^\ast$ and the trained Meta-Moderator under the same maximum debate budget.}
    \label{fig:debate_rounds}
\end{figure*}

\paragraph{Robustness to the number of debaters.}
Increasing the number of debaters from $N{=}2$ to $N{=}3$ while keeping the backbone and debate budget fixed (Table~\ref{tab:debaters_num}), Meta-Moderator remains competitive on most benchmarks, showing robustness beyond a specific debate size.

Overall, these results support moderation as a policy-like capability that transfers across tasks, heterogeneous system instantiations, and debate group sizes.

\subsection{Ablation: Decoupling Stopping and Adjudication}
To disentangle adaptive stopping from learned adjudication, we evaluate two decoupled variants under the same debaters, protocol, budget $T_{\max}$, and decoding: (i) \emph{Adjudication-only} fixes the debate length to 3 and uses Meta-Moderator only for the final answer; (ii) \emph{Stopping-only} uses Meta-Moderator for \texttt{CONTINUE}/\texttt{STOP} but applies a fixed heuristic aggregator upon stopping. Table~\ref{tab:stop_adj} shows that neither component alone consistently recovers the full model's gains, while combining adaptive stopping with learned adjudication performs best across all five benchmarks.

\subsection{Analysis of Debate Coverage and Adjudication}
\label{sec:gain_analysis}

To understand why Meta-Moderator helps, we decompose outcomes into \emph{coverage}: whether a correct hypothesis ever appears; \emph{adjudication}: whether it is selected once available.
For each instance, let $\mathcal{H}_t$ be the set of debaters' answers observed up to the moderator's stopping time $t$, and let $y^*$ denote the gold answer.
We count \emph{oracle availability} when the gold answer appears in the debate, $y^* \in \mathcal{H}_t$, and \emph{mis-aggregation} when the final prediction is still incorrect despite oracle availability, $\hat{y}\neq y^* \wedge y^* \in \mathcal{H}_t$.

As shown in Table~\ref{tab:oracle_single}, training does not uniformly increase oracle availability: coverage improves on AMC and MATH500, remains nearly unchanged on GSM8K, and decreases on StrategyQA and MMLU. In contrast, the trained Meta-Moderator consistently reduces mis-aggregation across all five benchmarks (from $33$ to $12$ on GSM8K, $8$ to $1$ on AMC, $25$ to $8$ on MATH500, $42$ to $9$ on StrategyQA, and $77$ to $15$ on MMLU). Since $\mathcal{H}_t$ is defined at the method-specific stopping time, lower coverage may partly reflect earlier termination rather than poorer hypothesis generation. These results suggest that the accuracy gains primarily arise from more reliable finalization when useful debate evidence is available, rather than from uniformly increasing debate coverage.

\subsection{Analysis of Debate Rounds}
We analyze stopping behavior by reporting the average number of debate rounds used by Meta-Moderator and the untrained Meta-Moderator$^\ast$ in Figure~\ref{fig:debate_rounds} (maximum budget $T_{\max}{=}5$).
Meta-Moderator$^\ast$ tends to over-deliberate, often consuming nearly the full budget, especially with the smaller backbone.
After training, Meta-Moderator uses substantially fewer rounds across benchmarks while remaining adaptive: it reduces redundant debate on easier instances but allocates more rounds when helpful on harder benchmarks (Figure~\ref{fig:debate_rounds}).

Overall, Meta-Moderator learns adaptive debate budgeting rather than a fixed stopping heuristic.

\section{Conclusion}
We identify moderation as a central bottleneck in multi-agent debate and argue that effective debate requires a meta-cognitive loop of monitoring, control, and adjudication.
We therefore propose \textbf{Meta-Moderator}, a learnable framework that regulates deliberation by deciding at each round whether to continue debating or to stop and commit to a final answer. The moderation policy is trained independently of debaters via outcome-driven reinforcement learning.
Across five reasoning benchmarks and two backbone settings, Meta-Moderator consistently improves accuracy over common debate decision layers and generalizes across tasks, heterogeneous moderator/debater backbones, and debate group sizes.
Analyses further show that it budgets debate adaptively, increases the likelihood that a correct hypothesis emerges, and reduces mis-aggregation once it does, enabling more reliable and efficient deliberation.

\section*{Limitations}
A key limitation of our approach is the additional training and inference cost of learning and running a dedicated moderator in conjunction with multi-agent debate.
Although Meta-Moderator often reduces debate rounds at test time, it still requires extra model calls for monitoring and adjudication, and outcome-driven RL training further increases compute relative to purely prompted decision layers.

Moreover, our moderator has a restricted control interface due to our data construction: we only consider the minimal \texttt{continue}/\texttt{stop} action space.
While this design isolates the effect of learnable stopping and adjudication, it does not capture richer interventions that may further improve debate quality.

Future work could reduce overhead by distilling the moderator to a smaller model, amortizing monitoring signals, or pruning agents adaptively.
It would also be valuable to expand the moderation action space and develop objectives and evaluation suites beyond final-answer accuracy, enabling moderators that learn not only \emph{when} to stop, but also \emph{how} to steer deliberation.

\section*{Acknowledgement}
This work was partially supported by NSFC/RGC Joint Research Scheme (N\_PolyU5179/25), Hong Kong RGC (PolyU25600624), Innovation Technology Fund (ITS/052/23MX, PRP/009/22FX), industrial sponsors and RMGS (P0045948, P0048183, P0048191, P0046453, P0060272). This work was also supported by the National Natural Science Foundation of China (Grant No.: 62372314). The experimental part of this work was supported by The Centre for Large AI Models (CLAIM) of The Hong Kong Polytechnic University.

\bibliography{custom}

\appendix
\clearpage

\section{More Implementation Details}
For large benchmarks, we evaluate all methods on the same fixed subset of 500 instances sampled once with a shared random seed, ensuring identical evaluation sets across methods. We use deterministic decoding (temperature=0) for all model components.

For inference, Qwen-3-30B-Instruct was run on two NVIDIA RTX A6000 GPUs, while all other models used one RTX A6000.
For GRPO training, the Llama-3.1-8B-Instruct, Llama-3.2-3B-Instruct, and Qwen-2.5-7B-Instruct models were each trained on one RTX A6000.

\section{Algorithm}
Algorithm~\ref{alg:meta_moderator_infer} describes our inference-time moderation protocol for multi-agent debate.

\begin{algorithm}[htbp]
\caption{Meta-Moderator Inference-Time Moderation for Multi-Agent Debate}
\label{alg:meta_moderator_infer}
\begin{algorithmic}[1]
\Require Input $x$, debaters $\mathcal{D}=\{d_1,\dots,d_N\}$, init operator $\mathrm{Init}$, debate operator $\mathrm{Debate}$, moderator policy $\pi_\theta$, state formatter $\phi(\cdot)$, max rounds $T_{\max}$.
\Ensure Final answer $\hat{y}$.

\State Initialize round $t \gets 0$
\State Initialize transcript $\mathcal{H}\gets\emptyset$

\For{$i=1$ to $N$}
    \State Initial response $r_{d_i}^{0} \sim \mathrm{Init}(x)$
\EndFor
\State Update transcript $\mathcal{H}\gets \mathcal{H}\cup \{r_{d_i}^{0}\}_{i=1}^{N}$

\While{$t \le T_{\max}$}
    \State Construct moderator state $s_t \gets \phi\big(x,\{r_{d_i}^{t}\}_{i=1}^{N}\big)$
    \State Moderator decision $z_t \sim \pi_{\theta}(\cdot \mid s_t)$
    \State Parse $(a_t,\hat{y}_t)\gets \mathrm{ParseDecision}(z_t)$
    \Comment{$z_t \in \{\texttt{CONTINUE},\ \texttt{STOP}\parallel \texttt{<answer>}\hat{y}_t\texttt{</answer>}\}$}

    \If{$a_t = \texttt{STOP}$}
        \State \textbf{return} $\hat{y}_t$
    \EndIf

    \If{$t = T_{\max}$}
        \Comment{No stop decision within budget; terminate with final moderator output}
        \State $\hat{y} \gets \mathrm{FinalizeAtBudget}(s_t,\pi_\theta)$
        \State \textbf{return} $\hat{y}$
    \EndIf

    \State \Comment{One-round parallel debate update}
    \State $t \gets t + 1$
    \For{$i=1$ to $N$ \textbf{in parallel}}
        \State Observe peers $\mathcal{R}^{t}_{-d_i} \gets \{r_{d_j}^{t-1}\mid j\neq i\}$
        \State Update response $r_{d_i}^{t} \sim \mathrm{Debate}\big(x;\mathcal{R}^{t}_{-d_i}\big)$
    \EndFor
    \State Update transcript $\mathcal{H}\gets \mathcal{H}\cup \{r_{d_i}^{t}\}_{i=1}^{N}$
\EndWhile

\end{algorithmic}
\end{algorithm}

\section{Learning Meta-Moderator with Reinforcement Learning}
\label{app:learning}

The GRPO update maximizes a PPO-style clipped surrogate objective with a KL regularizer:
\begin{equation}
\begin{aligned}
\mathcal{J}_{\text{GRPO}}(\theta)
&=
\mathbb{E}_{s_t \sim P(S),\, \{z_t^{(i)}\}_{i=1}^{G} \sim \pi_{\theta_{\text{old}}}(\cdot \mid s_t)} \\
&
\Bigg[
\frac{1}{G}\sum_{i=1}^{G}
\min\Big(
\rho_i(\theta)\,A^{(i)}, \\
& \operatorname{clip}\big(\rho_i(\theta), 1-\epsilon, 1+\epsilon\big)A^{(i)}
\Big) \\
&-\beta D_{\mathrm{KL}}\left(\pi_{\theta}(\cdot\mid s_t)\|\pi_{\mathrm{ref}}(\cdot\mid s_t)\right)
\Bigg]
\end{aligned}
\end{equation}

where $\rho_i(\theta) = \frac{\pi_{\theta}(z_t^{(i)}\mid s_t)}{\pi_{\theta_{\text{old}}}(z_t^{(i)}\mid s_t)}$
is the importance ratio. The clipping operator constrains the update magnitude within
$[1-\epsilon, 1+\epsilon]$ to stabilize learning, while the KL term regularizes the policy
toward a reference model $\pi_{\mathrm{ref}}$ with strength $\beta$.
In all experiments, we use a group size of $G{=}8$, a clipping ratio of $\epsilon{=}0.2$, and a KL coefficient of $\beta{=}0$, and optimize the moderator for $3000$ gradient-update steps.
This group-relative, clipped update yields stable policy improvements under sparse,
outcome-driven rewards and directly optimizes the moderator's regulation behavior.

\section{Prompt Templates}
\label{app:prompts}

This section presents the prompt templates used in our implementation for both the debaters and the Meta-Moderator.
We adopt a lightweight tag-based protocol for parsing: each debater ends with \texttt{<answer>}...\texttt{</answer>}, and the Meta-Moderator outputs either \texttt{CONTINUE} or \texttt{STOP} with an optional \texttt{<answer>} span.

\subsection{Prompt for Meta-Moderator (\texttt{STOP/CONTINUE})}
\label{app:prompt_moderator_stop_continue}

\begin{tcolorbox}[colback=black!1!white,colframe=black!57!white,title=Prompt for Meta-Moderator (STOP/CONTINUE), breakable]
You are a moderator in a debate. Your task is to decide whether to CONTINUE or STOP the debate based on the arguments presented by debaters.\\
If STOP, output the STOP first and then the final answer between <answer> and </answer> tags with no explanations or additional text.\\
If CONTINUE, output the CONTINUE only.\\
\\
Question:\\
\{question\}\\
\\
Other agents' responses:\\
\{agents\_responses\}\\
\end{tcolorbox}

\subsection{Prompt for Meta-Moderator (Forced Answer at $T_{\max}$)}
\label{app:prompt_moderator_forced_answer}

\begin{tcolorbox}[colback=black!1!white,colframe=black!57!white,title=Prompt for Meta-Moderator (Forced Answer at $T_{\max}$), breakable]
You are a moderator in a debate competition. Your task is to determine the correct final answer based on the arguments presented by debaters.\\
Output only the final answer with no explanations or additional text.\\
\\
Question:\\
\{question\}\\
\\
Other agents' responses:\\
\{agents\_responses\}\\
\end{tcolorbox}

\subsection{Debater Prompts}
\label{app:prompts_debater}

\begin{tcolorbox}[colback=black!1!white,colframe=black!57!white,title=Prompt for Debater Agent (Initial Round), breakable]
Answer the question based on your own knowledge. Explain your answer step by step, and put the answer between <answer> and </answer> tags at the end of your response.\\
\{optional confidence instruction\}\\
\\
Question:\\
\{question\}\\
\end{tcolorbox}

\begin{tcolorbox}[colback=black!1!white,colframe=black!57!white,title=Prompt for Debater Agent (Debate Round), breakable]
I will give the answers and arguments to this question from other agents. Use their solution as additional advice; note that they may be wrong. If you disagree with the other agents, please give your reasons and answer; otherwise, revise your previous answer. Explain your answer step by step, and put the answer between <answer> and </answer> tags at the end of your response.\\
\\
Question: \{question\}\\
Other agents' responses:\\
\{other\_agents\_responses\}\\
\end{tcolorbox}




\section{Training Template}
\label{app:training_template}

\paragraph{Training data generation.}
We construct the Meta-Moderator training set by first generating multi-agent debate traces using the standard Multi-Agent Debate (MAD) pipeline, with Llama3.1-8B-Instruct as the underlying generator.
For each question, MAD produces debater responses per round.
In our data construction, we only retain two debaters ($N=2$) and use their responses to form the debate state; any additional agent outputs are discarded.
Each training instance is a single-round snapshot consisting of the original question and the two agents' responses at that round, paired with a Meta-Moderator decision label (\texttt{CONTINUE} or \texttt{STOP}) and (when applicable) a target final answer.

\paragraph{Template.}
We train the Meta-Moderator using a single prompt template constructed from the current-round debate state.
The input consists of: (i) an instruction that constrains the output space to \texttt{CONTINUE} or \texttt{STOP}
(with \texttt{<answer>}...\texttt{</answer>} when stopping), and (ii) the question plus the two debaters' responses at the current round.

\begin{tcolorbox}[colback=black!1!white,colframe=black!57!white,title=Training Template for Meta-Moderator, breakable]
You are a moderator in a debate. Your task is to decide whether to CONTINUE or STOP the debate based on the arguments presented by debaters.
If STOP, output the STOP first and then the final answer between <answer> and </answer> tags with no explanations or additional text.
If CONTINUE, output the CONTINUE only.\\
\\
Question: \{question\}\\
Other agents' responses:\\
Agent 1: \{response\_1\}\\
Agent 2: \{response\_2\}\\
\end{tcolorbox}

\section{Efficiency and Cost Accounting}
We report the training and inference overhead of Meta-Moderator for completeness.

\subsection{Training cost}
Meta-Moderator requires constructing offline debate trajectories and performing RL optimization. The total generation cost for constructing training data is approximately 11.25M tokens (9.78M input + 1.46M output). The moderator is trained once and reused across tasks and backbone configurations; debaters are not fine-tuned.

\subsection{Inference cost, latency, and memory}
We additionally report inference token usage for all methods in Table~\ref{tab:cost_tokens} to quantify the accuracy–compute trade-off.
At inference time, all methods share the same debate protocol, backbone, maximum round budget ($T_{max}$), and decoding configuration; the only difference is the decision layer. Meta-Moderator introduces one additional moderator call per round. However, it typically reduces the average number of debate rounds (Figure~\ref {fig:debate_rounds}), thereby lowering the number of debater-side generation tokens. As a result, the net inference token cost is comparable and is often reduced due to adaptive early stopping. The memory overhead is limited to loading one additional moderator model, which can be smaller than the debaters (as demonstrated by the decoupled debater-moderator setting in Table~\ref{tab:generalization}). From an implementation perspective, our framework only replaces the decision layer and does not modify the debate protocol itself. Importantly, our claim is not that Meta-Moderator is universally cheaper than all prompt-based baselines, but that it improves the accuracy-compute trade-off by allocating deliberation adaptively.

\section{Trained Single-Agent GRPO Baseline}
\label{app:naive_grpo}
To isolate whether performance gains come from GRPO training alone, we implement a trained single-agent baseline in Table~\ref{tab:naive_grpo}.
We use the same backbone (Llama-3.1-8B-Instruct) and the same GRPO optimizer.
The reward is based on final-answer correctness with dataset-specific matching.
We train on the same training splits used to construct Meta-Moderator training data (500 instances each from GSM8K and MMLU).

Overall, Naive+GRPO can improve over naive prompting on some benchmarks, but it remains below Meta-Moderator.
This suggests that the gains are not explained by GRPO training alone, and that meta-level regulation of multi-agent debate (adaptive stopping and adjudication) provides additional benefits.

\begin{table*}[t]
    \centering
    \adjustbox{max width=\textwidth}{
    \begin{tabular}{lccccc}
        \toprule
        Method & GSM8K & AMC & MATH500 & StrategyQA & MMLU \\
        \midrule
        Naive & 22.20 & 7.23 & 3.80 & 68.20 & 57.00 \\
        Naive+GRPO & 81.60 & 19.28 & 4.60 & 68.80 & 62.80 \\
        \textbf{Meta-Moderator (Ours)} & \textbf{83.80} & \textbf{27.71} & \textbf{35.80} & \textbf{72.00} & \textbf{67.20} \\
        \bottomrule
    \end{tabular}
    }
    \caption{Trained single-agent GRPO baseline (Naive+GRPO) under Llama-3.1-8B-Instruct.}
    \label{tab:naive_grpo}
\end{table*}

\section{Additional Results with Different Backbone Settings}
\label{app:extra_results}

Shown in Table \ref{tab:main2}, we report additional backbone settings for the debate system beyond the main results.
We evaluate three settings:
(i) we use \textbf{Llama-3.2-3B-Instruct} \cite{dubey2024llama} as both the debaters and the moderator;
(ii) we use \textbf{Qwen-2.5-14B-Instruct} \cite{bai2023qwen} as the debaters, while using \textbf{Llama-3.1-8B-Instruct} as the moderator; and
(iii) we use \textbf{Qwen-3-30B-Instruct} \cite{bai2023qwen} as the debaters, while using \textbf{Llama-3.1-8B-Instruct} as the moderator.
Following our main setup, we construct the moderator state using responses from \textbf{two} debaters ($N=2$).
All numbers are accuracies (\%).

\begin{table*}[h]
    \centering
    \adjustbox{max width=\textwidth}{
    \begin{tabular}{lccccc} 
        \toprule
        Method & GSM8K & AMC & MATH500 & StrategyQA & MMLU \\ 
        \midrule
        \rowcolor{gray!15}\textbf{\textit{Single-agent}} & \multicolumn{5}{c}{\textit{Llama-3.1-8B-Instruct}} \\ 
        Naive & 147.89 & 210.96 & 252.53 & 87.38 & 175.95 \\
        Reflection & 2408.09 & 3573.96 & 3223.21 & 2118.11 & 2763.92 \\
        CoT & 336.06 & 614.43 & 516.29 & 354.07 & 485.06 \\
        \rowcolor{gray!15}\textbf{\textit{Multi-Agent Debate}} & \multicolumn{5}{c}{\textit{Llama-3.1-8B-Instruct}} \\ 
        + Majority-Voting & 10972.33 & 17785.76 & 16442.11 & 18460.87 & 13222.93 \\
        + Consensus & 10034.28 & 16890.30 & 15809.21 & 18740.09 & 12238.40 \\
        + LLM-as-Judge & 9401.76 & 15335.23 & 14399.28 & 14614.69 & 11084.35 \\
        + \textbf{Meta-Moderator$^*$} & 12329.51 & 8598.46 & 10060.58 & 7008.93 & 22350.50 \\
        + \textbf{Meta-Moderator} & 3522.08 & 12802.40 & 10668.56 & 4192.31 & 4017.78 \\
        \bottomrule
    \end{tabular}
    }
    \caption{Inference token statistics (average tokens per instance; input+output) across five benchmarks under the same debate protocol, backbone, round budget, and decoding configuration. Token counts include all LLM calls used by each method (e.g., debaters and the decision component).}
    \label{tab:cost_tokens}
\end{table*}

\begin{table*}[h]
    \centering
    \adjustbox{max width=\textwidth}{
    \begin{tabular}{lcccccc} 
        \toprule
        Method & Use Moderator & GSM8K & AMC & MATH500 & StrategyQA & MMLU \\
        \midrule
        \rowcolor{gray!15}\textbf{\textit{Single-agent}} & \multicolumn{6}{c}{\textit{Llama-3.2-3B-Instruct}} \\ 
        Naive & - & 4.80 & 3.61 & 2.60 & 66.00 & 32.80 \\
        Reflection & - & 17.80 & 2.41 & 16.20 & 47.00 & 39.00 \\
        CoT & - & 74.20 & 14.46 & 25.20 & 57.60 & 40.00 \\
        \rowcolor{gray!15}\textbf{\textit{Multi-Agent Debate}} & \multicolumn{6}{c}{\textit{Llama-3.2-3B-Instruct}} \\
        + Majority-Voting & \xmark & 28.80 & 3.61 & 13.60 & 44.80 & 32.40 \\
        + Consensus & \xmark & 33.40 & 2.41 & 16.20 & 47.00 & 39.00 \\
        + LLM-as-Judge & \cmark & 66.20 & 7.23 & 17.40 & 24.40 & 41.40 \\
        + \textbf{Meta-Moderator$^*$} & \cmark & 63.40 & 8.43 & 17.40 & 44.80 & 39.80 \\
        + \textbf{Meta-Moderator} & \cmark & 74.60 & 18.07 & 23.40 & 56.40 & 42.40 \\
        \midrule
        \rowcolor{gray!15}\textbf{\textit{Single-agent}} & \multicolumn{6}{c}{\textit{Qwen-2.5-14B-Instruct}} \\ 
        Naive & - & 28.80 & 9.64 & 7.20 & 69.80 & 67.40 \\
        Reflection & - & 92.40 & 33.73 & 47.60 & 77.60 & 67.00 \\
        CoT & - & 92.40 & 16.87 & 44.00 & 77.00 & 64.20 \\
        \rowcolor{gray!15}\textbf{\textit{Multi-Agent Debate}} & \multicolumn{6}{c}{\textbf{Debaters:} \textit{Qwen-2.5-14B-Instruct} + \textbf{Moderator: } \textit{Llama-3.1-8B-Instruct}} \\ 
        + Majority-Voting & \xmark & 92.00 & 44.58 & 55.00 & 77.60 & 67.80 \\
        + Consensus & \xmark & 92.00 & 44.58 & 55.60 & 77.40 & 68.20 \\
        + LLM-as-Judge & \cmark & 92.20 & 44.58 & 53.56 & 76.80 & 68.60 \\
        + \textbf{Meta-Moderator$^*$} & \cmark & 92.40 & 42.71 & 52.20 & 77.60 & 69.40 \\
        + \textbf{Meta-Moderator} & \cmark & 92.60 & 44.58 & 56.00 & 77.80 & 69.00 \\
        \midrule
        \rowcolor{gray!15}\textbf{\textit{Single-agent}} & \multicolumn{6}{c}{\textit{Qwen-3-30B-Instruct}} \\ 
        Naive & - & 52.20 & 13.25 & 14.20 & 73.00 & 78.40 \\
        Reflection & - & 93.00 & 7.23 & 33.00 & 73.00 & 76.60 \\
        CoT & - & 93.80 & 13.25 & 37.80 & 76.00 & 76.80 \\
        \rowcolor{gray!15}\textbf{\textit{Multi-Agent Debate}} & \multicolumn{6}{c}{\textbf{Debaters:} \textit{Qwen-3-30B-Instruct} + \textbf{Moderator: } \textit{Llama-3.1-8B-Instruct}} \\ 
        + Majority-Voting & \xmark & 93.00 & 27.71 & 30.40 & 73.40 & 75.20 \\
        + Consensus & \xmark & 93.40 & 24.10 & 28.20 & 69.80 & 68.60 \\
        + LLM-as-Judge & \cmark & 93.40 & 36.14 & 47.00 & 73.40 & 78.40 \\
        + \textbf{Meta-Moderator$^*$} & \cmark & 92.80 & 36.14 & 47.40 & 72.40 & 76.60 \\
        + \textbf{Meta-Moderator} & \cmark & 93.60 & 39.76 & 59.60 & 75.60 & 79.80 \\
        \bottomrule
    \end{tabular}
    }
    \caption{Additional results under different backbone choices for debaters and moderators.}
    \label{tab:main2}
\end{table*}
\paragraph{Discussion.}
Across both settings, using a learned Meta-Moderator consistently improves over heuristic aggregation baselines (Majority-Voting / Consensus) and is competitive with (or better than) an LLM-as-Judge.
Notably, in the 7B setting, Meta-Moderator yields clear gains on MATH500 and StrategyQA compared to heuristic aggregation, indicating better debate termination and answer selection.

\section{Case Study}
\label{sec:case_study}

\begin{figure*}[t]
    \centering
    \includegraphics[width=0.98\linewidth]{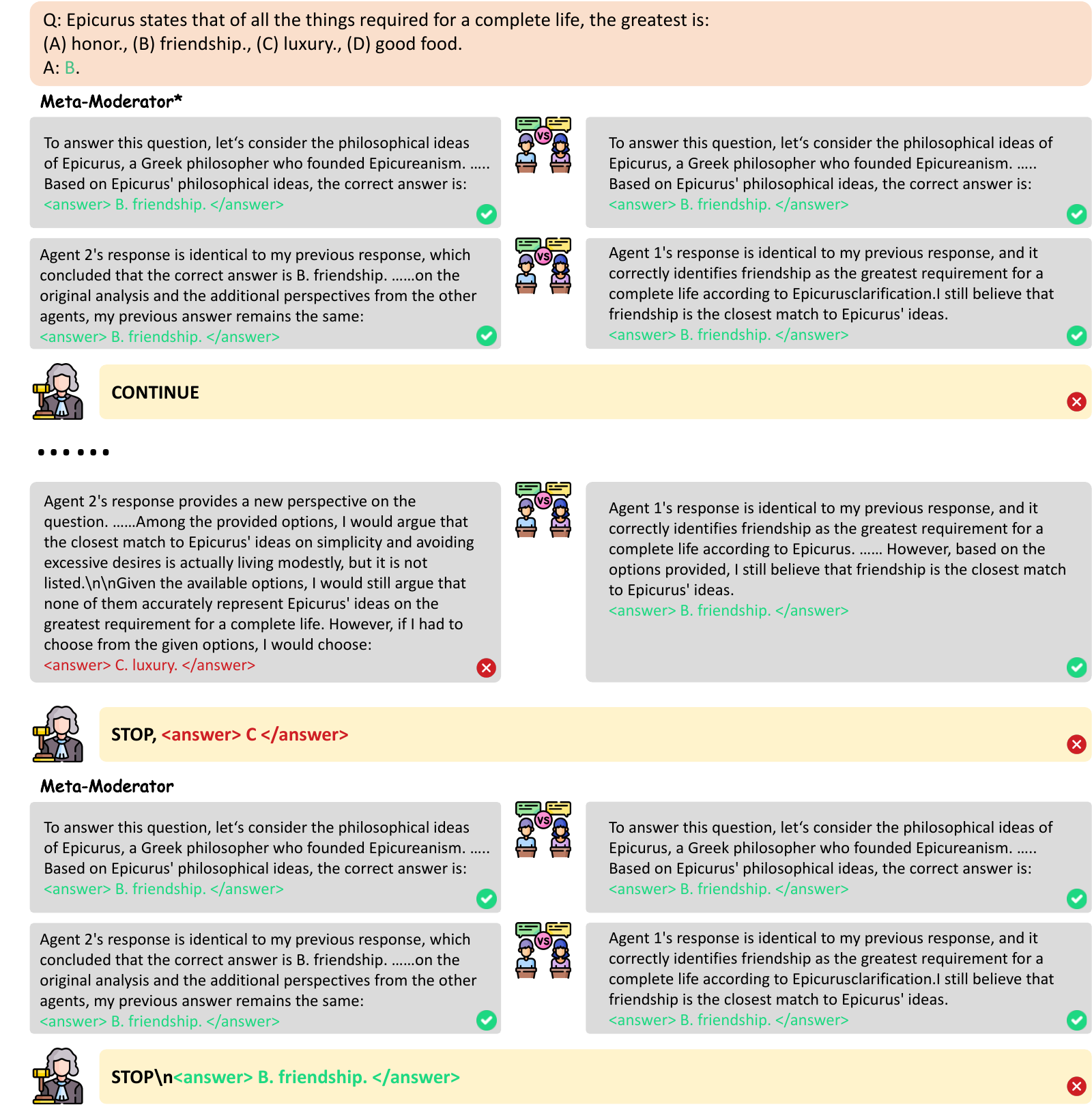}
    \caption{Case study comparing a failure of the untrained Meta-Moderator$^*$ with a successful decision by the trained Meta-Moderator.}
    \label{fig:case_study1}
\end{figure*}

We present a qualitative case study illustrating how moderation failures can negate the potential benefits of multi-agent debate, and how a trained moderator mitigates these issues.
Figure~\ref{fig:case_study1} shows an instance where the untrained \textbf{Meta-Moderator$^\ast$} fails to stop in time and eventually commits to an incorrect answer due to late-round drift, whereas the trained \textbf{Meta-Moderator} stops earlier and selects the correct hypothesis.

\paragraph{Oracle availability but mis-aggregation.}
In the early rounds, both debaters independently propose the correct option (\texttt{B. friendship}), meaning the gold hypothesis is already present in the debate history (i.e., oracle availability holds, $y^* \in \mathcal{H}_t$).
However, \textbf{Meta-Moderator$^\ast$} chooses to \texttt{continue}, spending additional rounds despite diminishing marginal utility.
As the debate proceeds, one agent introduces a spurious line of reasoning (arguing that none of the options perfectly match and then selecting \texttt{C. luxury} by elimination), which shifts the discussion away from the previously consistent correct hypothesis.
The moderator then \texttt{stops} and commits to the wrong answer, constituting a mis-aggregation error despite oracle availability ($\hat{y} \neq y^* \wedge y^* \in \mathcal{H}_t$).

\paragraph{Why learned moderation helps.}
This example highlights two failure modes of untrained moderation: (i) \emph{over-deliberation}, where redundant interaction consumes budget after a correct hypothesis has emerged, and (ii) \emph{problem drift}, where additional rounds introduce distracting arguments that increase the risk of mis-aggregation.
In contrast, the trained Meta-Moderator recognizes early decision readiness and terminates the debate before drift dominates, selecting \texttt{B. friendship}.
This behavior is consistent with our quantitative findings that learning a moderation policy reduces unnecessary rounds and lowers mis-aggregation once informative hypotheses appear (Sections~\ref{sec:gain_analysis} and Figure~\ref{fig:debate_rounds}).

\end{document}